# VEHICLE PRICE PREDICTION BY AGGREGATING DECISION TREE MODEL WITH BOOSTING MODEL

Auwal Tijjani Amshi,

auwalamshi@gmail.com

**Abstract**

**Predicting the price for used vehicles is more interesting and needed problem by many users. Vehicle price prediction can be a challenging task due to the high number of attributes that should be considered for the accurate prediction. The major step in the prediction process is collection and pre-processing of the data. In this project, python scripts were built to normalize, standardize and clean data to avoid unnecessary noise for machine learning algorithms. The data set used in this project can be very valuable in conducting similar research using different prediction techniques. Many assumptions were made on the basis of the data set. The proposed system uses Decision tree model and Gradient boosting predictive model, which are combine in other to get closed to accurate prediction, the proposed model was evaluated and it gives a promising performance. The future price prediction of used vehicles with the help of same data set will comprise different model.**

## Introduction

### 1.1 Background

Predicting the price for used vehicles is more interesting and needed problem by many users. The data set has been collected in the year of 2014 from the statistics field says that 84 percent of vehicles had been bought for their own personal usage (Agencija et al, 2018). Later this percent has been increased very much because people has more interested to buy new vehicles often and also showing more interest on finding the price of used vehicles. This increases the efficiency of prediction techniques. Only with the help of experts and their corresponding knowledge we can achieve more accurate prediction, for used vehicle price. So the varying prediction algorithms from machine learning suit this topic more efficiently. While predicting price of used vehicles we need an entire different features and factors. The most significant feature is brand and model of vehicle and also the mileage plays the major role for predicting the price of used vehicle. The most popular ingredient for vehicle is type of fuel and the volume of fuel in which it consumes for each mile. This particular data set might highly affect the price of a vehicle. And also we need to consider the price of fuel because it may changes frequently. The variety of features such as exterior color, door number, type of transmission, dimensions, safety, air condition, interior, whether it has navigation or not will also influence the vehicle price.

Many researchers proposed machine learning algorithm which all have their purposes but in this project, we will be using machine learning algorithm named "Decision Tree" to design a system that can be able to the prediction of used vehicle price.

### 1.2 Aims and Objectives

This research work aims at designing and implementing used Vehicle price prediction using machine learning algorithm. Which will address issues mentioned below?

1) Design a model to predicted price of used vehicle.
2) To Aggregating Gradient Boosting with decision tree to produce accurate predict.
3) To evaluate the usability of the propose application.

### 1.3 Significance of the Study

Price prediction is somehow interesting and popular problem. Accurate vehicle price prediction involves expert knowledge, because price usually depends on many distinctive features and factors. Typically, most significant ones are brand and model, age, horsepower and mileage. The fuel type used in the Vehicle as well as fuel consumption per mile highly affect price of a Vehicle due to a frequent changes in the price of a fuel. Different features like exterior color, door number, type of transmission, dimensions, safety, air condition, interior, whether it has navigation or not will also influence the vehicle price.

Deciding whether a used vehicle is worth the posted price when you see listings online can be difficult. Several factors, discuss above can influence the actual worth of a vehicle. From the perspective of a seller, it is also a dilemma to price a used Vehicle appropriately. Based on existing data, the aim is to use machine learning algorithms to design and develop system for predicting vehicle prices.

## 1.4 Definition of Terms

- Machine learning

Machine learning is the study of computer algorithms that improve automatically through experience and by the use of data. It is seen as a part of artificial intelligence.

- Algorithm

In computer science, programming, and math, an algorithm is a sequence of instructions where the main goal is to solve a specific problem, perform a certain action, or computation. In some way, an algorithm is a very clear specification for processing data, for doing calculations, among many other tasks.

- Predictive algorithms

Predictive algorithms (modeling) is a statistical technique using machine learning and data mining to predict and forecast likely future outcomes with the aid of historical and existing data. It works by analyzing current and historical data and projecting what it learns on a model generated to forecast likely outcomes.

- Data mining

Data mining is the process of analyzing a large batch of information to discern trends and patterns.

- Decision tree

Decision tree methodology is a commonly used data mining method for establishing classification systems based on multiple covariates or for developing prediction algorithms for a target variable.

**Literature Review**

## 2.1 Introduction

The system is designed to predict price of used vehicles and hence appropriate algorithms must be used to do the task. Before the algorithms are selected for further use, we conduct a literature review on several machine learning, prediction models and evaluation models.

## 2.2 Machine Learning

Learning has been described by Simon as the process of improving behavior through the discovery of new information over time. The learning is called Machine learning when perform by a machine. The concept of improvement is the status of finding the best solution for future problems by gaining experience from the existing examples in the process of machine learning (Sirmacek, B. 2007). With the development of information technologies over time, the concept of big data has emerged. The concept of big data is defined as very large and raw data sets that limitless and continues to accumulate, which cannot be solved by traditional databases methods (Altunisik, R. 2015).

The operations performed on the computer using the algorithm are performed according to a certain order without any margin of error. However, unlike the commands created to obtain the output from the data entered in this way, there are also cases where the decision making process takes place based on the sample data already available. In such cases, computers can make the wrong decisions such as mistakes that people can make in the decision-making process. In other words, machine learning is to gain a learning ability similar to human brain to computer by taking advantage of data and experience (Gor, I. 2014). The primary aim of machine learning is to develop models that can train to develop themselves and by detecting complex patterns and to create models to solve new problems based on historical data (Turkmenoglu, C. 2016).

Machine learning and data-driven approaches are becoming very important in many areas. For example, smart spam classifiers protect our e-mails by learning from large amounts of spam data and user feedback. Ad systems learn to match the right ads with the right content; fraud detection systems protect banks from malicious attackers; Anomaly event detection systems help experimental physicists to find events that lead to new physics.

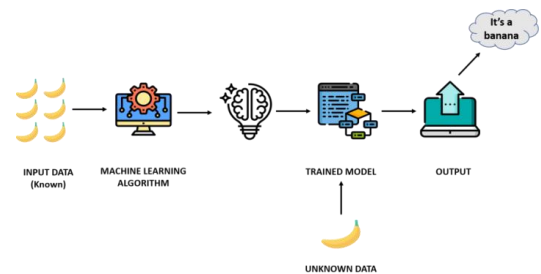

*Figure 2.1:    Classification Machine learning model.*

Figure 2.1 illustrate a conceptual design of a banana classification model, where the model should accurately to classify if the date presented/inputted is for a banana or not ("it's a banana" or "it's not a banana")

## 2.3 Prediction models

We have several prediction models in machine learning, this project selected two models which are gradient boosting and decision Tree Regression, which will be discuss in this section.

### 2.3.1 Decision Tree Regression

A decision tree is a classifier (machine learning algorithm) expressed as a recursive partition of the in-stance space. The decision tree consists of nodes that form a rooted tree, meaning it is a directed tree with a node called "root" that has no in coming edges. All other nodes have exactly one incoming edge. A node with outgoing edges is called an internal or test node. All other nodes are called leaves (also known as terminal or decision nodes). In a decision tree, each internal node splits the instance space into two or more sub-spaces according to a certain discrete function of the input attributes values. In the simplest and most frequent case, each test considers a single attribute, such that the instance space is partitioned according to the attribute's value. In the case of numeric attributes, the condition refers to a range. Example of a decision tree is shown in figure 2.2

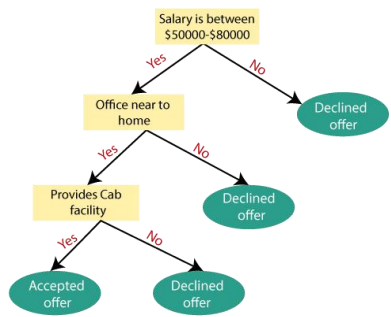

*Figure 2.2: An example for Decision Tree Model*

### 2.3.2 Gradient boosting

Gradient boosting is a machine learning ensemble meta-algorithm for primarily reducing bias, and also variance in supervised learning, and a family of machine learning algorithms which convert weak learners to strong ones. Boosting is based on the question posed by Kearns and Valiant (1988, 1989): Can a set of weak learners create a single strong learner? A weak learner is defined to be a classifier which is only slightly correlated with the true classification (it can label examples better than random guessing). In contrast, a strong learner is a classifier that is arbitrarily well-correlated with the true classification. (~Wikipedia)

Gradient Boosting relies on the intuition that the best possible next model, when combined with the previous models, minimizes the overall prediction errors. The key idea is to set the target outcomes from the previous models to the next model in order to minimize the errors. This is another boosting algorithm (few others are gradient boosting, Adaboost, XGBoost etc.).

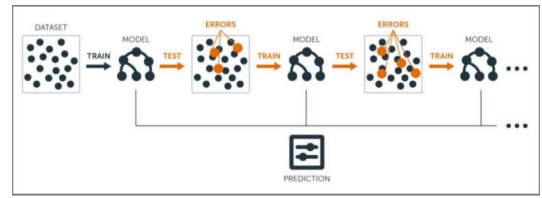

*Figure 2.3: An example for Gradient Boosting Model*

## 2.4 Evaluation Models

This project selected two well known evaluating models which are root-mean-square deviation (RMSD) and variance score, which will be discuss in below.

### 2.4.1 Root Mean Square Deviation (RMSD)

The root-mean-square deviation (RMSD) or root-mean-square error (RMSE) (or sometimes root-mean-squared error) is a frequently used measure of the differences between values (sample and population values) predicted by a model or an estimator and the values actually observed. The RMSD represents the sample standard deviation of the differences between predicted values and observed values. These individual differences are called residuals when the calculations are performed over the data sample that was used for estimation, and are called prediction errors when computed out-of-sample. The RMSD serves to aggregate the magnitudes of the errors in predictions for various times into a single measure of predictive power. RMSD is a measure of accuracy, to compare forecasting errors of different models for a particular data and not between datasets, as it is scale-dependent. (~Wikipedia)

$$\text{RMSD} = \sqrt{\frac{\sum_{i=1}^{N}(x_i - \hat{x}_i)^2}{N}}$$

(2.1)

Where RMSD is root-mean-square deviation, $N$ is number of non missing data points, $\hat{x}$ is the estimated value, $x$ is the actual value.

### 2.4.1 Variance Score ($r^2 score$)

Variance Score is "(total variance explained by model) / total variance." So if it is 100%, the two variables are perfectly correlated, i.e., with no variance at all. A low value would show a low level of correlation, meaning a regression model that is not valid, but not in all cases. (~Wikipedia)

$$r^2 score = \frac{\text{Total variance explained by model}}{\text{Total variance.}}$$
(2.2)

## 2.5   Related Work

Researchers more often predict prices of products using some previous data and so did Pudaruth (Pudaruth, S. 2014) who predicted prices of cars in Mauritius and these cars were not new rather second hand. He used multiple linear regression, k-nearest neighbors, naïve Bayes and decision trees algorithm in order to predict the prices. The comparison of prediction results from these techniques showed that the prices from these methods are closely comparable. However, it was found that decision tree algorithm and naïve bayes method were unable to classify and predict numeric values. Pudaruth's research also concluded that limited number of instances in data set do not offer high prediction accuracies (Pudaruth, S. 2014).

Multivariate regression model helps in classifying and predicting values of numeric format. Kuiper, 2008 used this model to predict price of 2005 General Motor (GM) cars. The price prediction of cars does not require any special knowledge so the data available online is enough to predict prices like the data available on www.pakwheels.com. Kuiper, 2008 did the same i.e. car price prediction and introduced variable selection techniques which helped in finding which variables are more relevant for inclusion in model. He encouraged students to use different models and find how checking model assumptions work.

Another similar research by (Listiani, 2009) uses Support Vector Machines (SVM) to predict the prices of leased cars. This research showed that SVM is far more accurate in predicting prices as compared to the multiple linear regression when a very large dataset is available. SVM also handles high dimensional data better and avoids both the under-fitting and over-fitting issues. Genetic algorithm is used by Listiani (2009) to find important features for SVM. However, the technique does not show in terms of variance and mean standard deviation why SVM is better than simple multiple regressions.

Jain, Yash. (2019). Used a simple linear regression using python to perform price prediction. The data used in the report is completely hypothetical.

Awoke, T., et al. (2021), target is to implement the efficient deep learning-based prediction models specifically long short-term memory (LSTM) and gated recurrent unit (GRU) to handle the price volatility of bitcoin and to obtain high accuracy. The study involves comparing these two time series deep learning techniques and proved the efficacy in forecasting the price of bitcoin.

## 2.6   Summary of Related Studies

In this section we summaries the related studies discuss in related work section.

*Table 2.1:   Summary of Related Studies*

| S/N | Author/Year | Title/method/techniques | Limitations/Findings |
|---|---|---|---|
| 1 | Pudaruth, S. 2014 | Multiple Linear Regression, K-Nearest Neighbors, Naïve Bayes And Decision Trees Algorithm. | The research has limited number of instances in data set does not offer high prediction accuracies. |
| 2 | Aaron Ng, 2015 | Support Vector Machines (SVM) | The technique does not show in terms of variance and mean standard deviation why SVM is better than simple multiple regressions. |
| 3 | Jain, Yash. (2019). | Simple Linear Regression | The data used in the report is completely hypothetical. |
| 4 | Awoke, T., et al. (2021) | long short-term memory (LSTM) and gated recurrent unit (GRU) | The study involves comparing these two time series deep learning techniques and proved the efficacy in forecasting the price of bitcoin. |

## 3.1 Methodology

The project methodology is a careful study or investigation, especially in order to discover new knowledge or information. One of the important issues in any machine learning project is deciding what programming language to be used. Python programming is among the common and popular development technology, which will be discus further in this section.

We divided the processes in this project into two phases, which is illustrated in figure 3.1

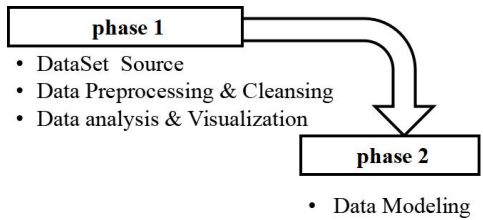

*Figure 3.1: Project Phase.*

The figure above illustrate the three phase followed in other to complete this project,

Phase 1, which include data sourcing, data pre-processing and cleaning. After sourcing data and cleaning the data we analyzes and visualize the data in other to get a unique inside and the relationship data points in the data. Phase 2, there are two primary steps under data modeling in the project:

1. **Training phase**: The system is trained by using the data in the data set and fits a model (line/curve) based on the algorithm chosen accordingly.

2. **Testing phase**: the system is provided with the inputs and is tested for its working. The accuracy is checked. And therefore, the data that is used to train the model or test it has to be appropriate.

The model is designed to predict price of used vehicle and hence appropriate algorithms must be used to do the two different tasks. Before the algorithms are selected for further use, different algorithms were compared for its accuracy. The well-suited one for the task was chosen.

## 3.2 Dataset Source

The data of this project was extracted from a public Online Vehicle selling website and an open source dataset site "https://www.kaggle.com/"

After getting the RAW data, the next step is to preprocess these dataset in order to make it useful for visualization and training session. First of all we read our dataset into a DataFrame (Table consists of rows and columns), so that we can manipulate it easily. A sample data table is shown in Table 3.1.

*Table 3.1: Sample Data*

|   | name | year | selling_price | km_driven | fuel | seller_type | transmission | owner | mileage | engine | max_power | torque | seats |
|---|---|---|---|---|---|---|---|---|---|---|---|---|---|
| 0 | Maruti Swift Dzire VDI | 2014 | 450000 | 145500 | Diesel | Individual | Manual | First Owner | 23.4 kmpl | 1248 CC | 74 bhp | 190Nm@ 2000rpm | 5.0 |
| 1 | Skoda Rapid 1.5 TDI Ambition | 2014 | 370000 | 120000 | Diesel | Individual | Manual | Second Owner | 21.14 kmpl | 1498 CC | 103.52 bhp | 250Nm@ 1500-2500rpm | 5.0 |
| 2 | Honda City 2017-2020 EXi | 2006 | 158000 | 140000 | Petrol | Individual | Manual | Third Owner | 17.7 kmpl | 1497 CC | 78 bhp | 12.7@ 2,700(kgm@ rpm) | 5.0 |
| 3 | Hyundai i20 Sportz Diesel | 2010 | 225000 | 127000 | Diesel | Individual | Manual | First Owner | 23.0 kmpl | 1396 CC | 90 bhp | 22.4 kgm at 1750-2750rpm | 5.0 |
| 4 | Maruti Swift VXI BSIII | 2007 | 130000 | 120000 | Petrol | Individual | Manual | First Owner | 16.1 kmpl | 1298 CC | 88.2 bhp | 11.5@ 4,500(kgm@ rpm) | 5.0 |

## 3.3 Data Preprocessing & Cleansing

The first thing I have to do is to clean unwanted strings from my columns, then change it to the appropriate type since all of them are string values and finally drop unwanted information (columns) from the data.

*Table 3.2: Sample data after preprocessing & cleansing*

|   | name | year | selling_price | km_driven | fuel | seller_type | transmission | owner | mileage | engine | max_power | seats |
|---|---|---|---|---|---|---|---|---|---|---|---|---|
| 0 | Maruti Swift Dzire VDI | 2014 | 450000 | 145500 | Diesel | Individual | Manual | First Owner | 23.40 | 1248 | 74.00 | 5 |
| 1 | Skoda Rapid 1.5 TDI Ambition | 2014 | 370000 | 120000 | Diesel | Individual | Manual | Second Owner | 21.14 | 1498 | 103.52 | 5 |
| 2 | Honda City 2017-2020 EXi | 2006 | 158000 | 140000 | Petrol | Individual | Manual | Third Owner | 17.70 | 1497 | 78.00 | 5 |
| 3 | Hyundai i20 Sportz Diesel | 2010 | 225000 | 127000 | Diesel | Individual | Manual | First Owner | 23.00 | 1396 | 90.00 | 5 |
| 4 | Maruti Swift VXI BSIII | 2007 | 130000 | 120000 | Petrol | Individual | Manual | First Owner | 16.10 | 1298 | 88.20 | 5 |

As shown in figure 3.2, the difference is clear when compare with the sample data illustrated in figure 3.1.

### 3.3.1 Dealing with Categorical Features

Since, we still have 3 categorical features (column) which are the fuel, seller_type, transmission owner and model the aim of this section is to pre process those features in order to make them numerical so that they will fit into our model.

In literature there is two famous kind of categorical variable transformation, the first one is **label encoding**, and the second one is the **one hot encoding**, for this use case we will use the one hot position and the reason why we choose this kind of data labeling is because we will not need any kind of data normalization later, and also This has the benefit of not weighting a value improperly but does have the downside of adding more columns to the data set.

### 3.3.2 Data Splitting

Usually we split our data into three parts: Training, validation and Testing set, but for simplicity we will use only train and test with 20% in test size.

## 3.4 Exploratory data analysis & Visualization

In other to understand the dataset generated we need to illustrate the relationship between the key data points such as the visualize relationship between car price and year model in Figure 3.1

### 3.4.1 Price distribution by year model

In this section we visualize the distribution of vehicles price by their year model release, and look how it behaves.

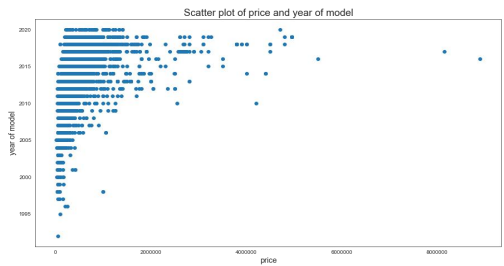

*Figure 3.2: Price distribution by year model*

As we can see from the figure 3.2 above, the vehicles price increase respectively by years, and more explicitly we can say that the more the car is recently released, the price augment, while in the other side the oldest vehicles still have a low price, and this is totally logical since whenever the cars become kind of old from the date of release, so their price start decrease.

### 3.4.2 Price distribution by KM Driven

In this section we visualize the distribution of cars price by their KM Driven, and look how it behaves.

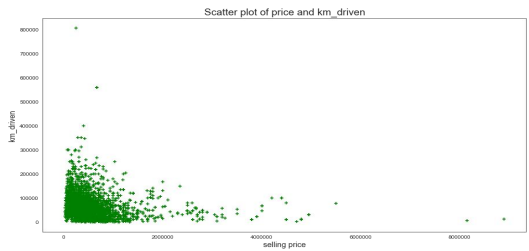

*Figure 3.3: Price distribution by KM Driven*

Form figure 3.3 we can see that as the used vehicles have Kilometer driven is increasing the price decreases. Which tells us that, the vehicles that have less km-driven have high price.

### 3.4.3 Correlation Matrix of the Data points in the Data Set

Although weak, it appears that there seems to be a positive relationship. Let's find the actual correlation between price and the other data points.

We will look at this in 2 ways heatman for visualization and the correlation coefficient score.

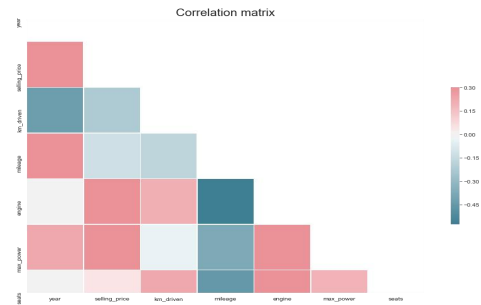

*Figure 3.4: Correlation Matrix of the dataset*

## 3.5 Data modeling (proposed model)

This project combines decision tree and gradient boosting to get close to accurate result.

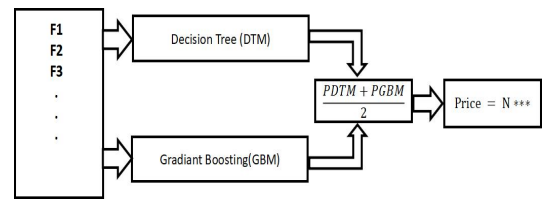

*Figure 3.5: Proposed Model Architecture*

For the proposed model architecture, (F2, F2, F3…) represent the features provided as input of the machine learning model (GNM) used in this project, each of the machine learning models will be used to predict a vehicle price then finally we add and get the average of the two predicted price, which will be the final price predicted. Moreover each of the selected predictive model used in this project are proven as good predictive model in chapter 4.

## 3.6 Choose of programming Language

The programming language use for this project work is Python. Python is an interpreted high-level general-purpose programming language. Python's design philosophy emphasizes code readability with its notable use of significant indentation. Its language constructs as well as its object-oriented approach aim to help programmers write clear, logical code for small and large-scale projects (Kuhlman and Dave. 2012).

## 4.1 Experiment Result

In this chapter we perform several experiments our dataset and evaluate the proposed system by comparing the performance of the proposed model with other prediction model such as KNN, Decision Tree etc.

The standardized residual is a measure of the strength of the difference between observed and expected values.)

### 4.1.1 Experiment 1: Vehicle price prediction using Decision Tree

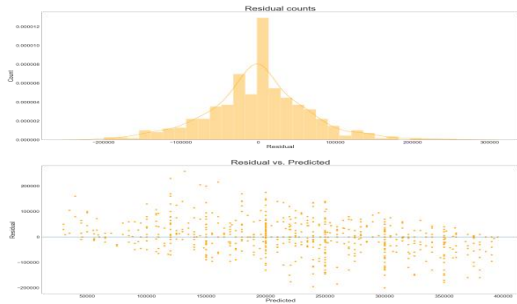

Figure 4.1: *Decision Tree Model*

Figure 4.1 illustrate the difference between the real price and the predicted price using Decision Tree Model. Furthermore we used a simple line graph in figure 4.2 to show the difference between the Real price vehicle and predicted vehicle price by Decision Tree Model.

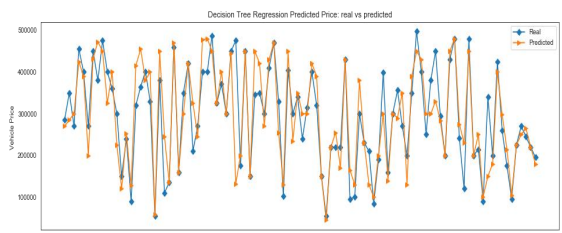

Figure 4.2: *Real price and Predicted price by DTM.*

### 4.1.2 Experiment 2: Vehicle price prediction using gradient boosting

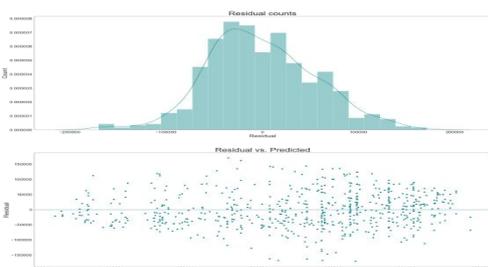

Figure 4.3: *gradient boosting Model*

Figure 4.3 illustrate the difference between the real price and the predicted price using gradient boosting Model. Figure 4.4 to show the difference between the Real price vehicle and predicted vehicle price by gradient Boosting Model.

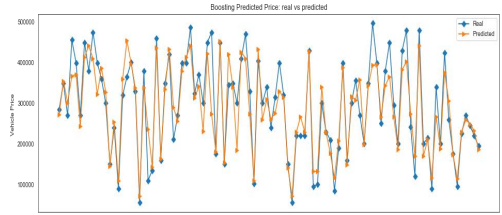

Figure 4.4: *Real price and Predicted price by GBM.*

### 4.1.3 Experiment 3: Model Evaluation.

To evaluate the proposed system we compare the RMSE Value and Variance Score ($r^2 score$) of several machine learning prediction model with proposed model. The result is shown in Table 4.1.

I. *Prediction Accuracy Results*

| Models | RMSE Value | Variance Score ($r^2 score$) |
|---|---|---|
| KNN Regression | 97311.22 | 0.30 |
| Linear Regression | 69139.25 | 0.65 |
| Decision Tree Regression | 65617.65 | 0.68 |
| Gradient Boosting | 51141.02 | 0.81 |

Table 4.2 shows first 10 Price predictions made by the proposed model which combine decision tree and gradient boosting model compare with the real price.

II. *Real price and predicted prices by the proposed model.*

| Real price | Gradient Boosting Price | Decision Tree Price | Propose Model Price |
|---|---|---|---|
| 3,000,000.00 | 2,928,770.16 | 3,000,000.00 | 2,964,385.08 |
| 315,040.00 | 601,930.41 | 400,000.00 | 500,965.21 |
| 1,100,000.00 | 1,334,670.50 | 1,100,000.00 | 1,217,335.25 |
| 1,950,000.00 | 1,681,160.57 | 2,500,000.00 | 2,090,580.29 |
| 3,400,000.00 | 3,568,470.44 | 3,500,000.00 | 3,534,235.22 |
| 3,300,000.00 | 2,872,000.29 | 3,800,000.00 | 3,336,000.15 |
| 4,900,000.00 | 4,684,360.03 | 4,900,000.00 | 4,792,180.02 |
| 2,500,000.00 | 2,635,350.43 | 2,400,000.00 | 2,517,675.22 |
| 2,900,000.00 | 3,187,440.33 | 4,950,000.00 | 4,068,720.17 |
| 800,000.00 | 1,075,570.39 | 1,000,000.00 | 1,037,785.20 |

## 4.2 Discussion

From Figure 4.1 which illustrates the standardized residual vehicle price prediction using decision Tree, we can conclude that the decision Tree can be used to accurately predict vehicle price prediction.

In Figure 4.2 we compare the Real price of vehicle and Predicted price by Decision tree model and from the graph we can conclude that decision tree model is a good predictive model. In addition, Table 4.1 shows the result Mean square error (56685.02) and variance score (0.63) of decision tree model, variance score can be considered as the prediction accuracy of decision tree model.

From Figure 4.3 which illustrates the standardized residual vehicle price prediction using gradient boosting, we can conclude that the gradient boosting can be used to accurately predict vehicle price prediction. Moreover in Figure 4.4 we compare the Real price of vehicle and Predicted price by boosting model and from the graph we can conclude that decision tree model is a good predictive model. In addition,

Table 4.1 shows the result Mean square error (45001.34) and variance score (0.77) of gradient boosting model, which has the higher accuracy and lowest mean square error when compare with KNN, Linear Regression ,Decision Tree.

Table 4.2 further illustrate the first 10 predicted price and compare with the real price and conclude that the proposed model did a good job in predicting an estimated price of used vehicle.

### 5.1 Summary and Conclusions

Vehicle price prediction can be a challenging task due to the high number of attributes that should be considered for the accurate prediction. The major step in the prediction process is collection and pre-processing of the data. In this project, python scripts were built to normalize, standardize and clean data to avoid unnecessary noise for machine learning algorithms.

The data set used in this project can be very valuable in conducting similar research using different prediction techniques. The prices of vehicles can be predicted using this data set on same or different prediction software as well. The data obtained under this research facilitated in prediction of prices of used vehicles through gradient boosting and decision tree regression method. Many assumptions were made on the basis of the data set. The proposed system was evaluated and it gives a promising performance. The future price prediction of used vehicles with the help of same data set will comprise different model.

### 5.2 Future Work

Although, this system has achieved astonishing performance in car price prediction problem our aim for the future research is to test this system to work successfully with various data sets. We will extend our test data with eBay (kaggle, 2018) and OLX (OLX 2018) used cars data sets and validate the proposed approach.